\documentclass[letterpaper, 10 pt, journal, twoside]{IEEEtran}
\usepackage{hyperref, url, amsmath, amssymb, booktabs, graphicx, bm, color, algorithm, tabularx, enumitem, multirow, threeparttable, flushend}
\usepackage[noend]{algpseudocode}
\usepackage{amsfonts}
\usepackage{booktabs} % Include this in the preamble for cleaner tables
\usepackage{multirow} % Include this for multirow support
\usepackage{makecell} % Include this for line breaks inside cells
\usepackage{algpseudocode}
\newcommand{\vg}[1]{\bm{#1}}
\renewcommand{\v}[1]{\mathbf{#1}}
\hypersetup{
    colorlinks=true,
    linkcolor=blue,
    citecolor=blue,
    filecolor=magenta,      
    urlcolor=cyan,
    pdftitle={Overleaf Example},
    pdfpagemode=FullScreen,
    }
\graphicspath{{images/}}
\usepackage{soul}

\newcolumntype{Y}{>{\centering\arraybackslash}X}

\makeatletter

\newcommand{\Rmnum}[1]{\expandafter\@slowromancap\romannumeral #1@}
\makeatother

\begin{document}
\bstctlcite{IEEEexample:BSTcontrol} % Enable the IEEEtranBSTCTL setting in .bib

\title{\LARGE Opt2Skill: Imitating Dynamically-feasible Whole-Body Trajectories\\ for Versatile Humanoid Loco-Manipulation}

% \author{Anonymous Author(s)}
\author{Fukang Liu, Zhaoyuan Gu, Yilin Cai, Ziyi Zhou, Hyunyoung Jung, Jaehwi Jang, Shijie Zhao,\\ Sehoon Ha, Yue Chen, Danfei Xu, and Ye Zhao % <-this % stops a space
% \thanks{Manuscript received: June, 15, 2025; Revised August, 25, 2025; Accepted September, 24, 2025.}
% \thanks{This paper was recommended for publication by Editor Abderrahmane Kheddar upon evaluation of the Associate Editor and Reviewers' comments.
% This work was supported by USDA-NIFA’s National Robotics Initiative Award 2022-11065 in collaboration with the National Science Foundation. \textit{(Corresponding author: Ye Zhao.)}}
\thanks{Authors are with the Institute for Robotics and Intelligent Machines, Georgia Institute of Technology, Atlanta, GA, USA.}
% \thanks{Digital Object Identifier (DOI): see top of this page.}
}

\markboth{IEEE Robotics and Automation Letters. Preprint Version. Accepted SEPTEMBER, 2025}
{Liu \MakeLowercase{\textit{et al.}}: Opt2Skill: Imitating Dynamically-feasible Whole-Body Trajectories for Versatile Humanoid Loco-Manipulation}

\maketitle

\begin{abstract}
Humanoid robots are designed to perform diverse loco-manipulation tasks. However, they face challenges due to their high-dimensional and unstable dynamics, as well as the complex contact-rich nature of the tasks. Model-based optimal control methods offer flexibility to define precise motion but are limited by high computational complexity and accurate contact sensing. On the other hand, reinforcement learning (RL) handles high-dimensional spaces with strong robustness but suffers from inefficient learning, unnatural motion, and sim-to-real gaps. To address these challenges, we introduce Opt2Skill, an end-to-end pipeline that combines model-based trajectory optimization with RL to achieve robust whole-body loco-manipulation. Opt2Skill generates dynamic feasible and contact-consistent reference motions for the Digit humanoid robot using differential dynamic programming (DDP) and trains RL policies to track these optimal trajectories. Our results demonstrate that Opt2Skill outperforms baselines that rely on human demonstrations and inverse kinematics-based references, both in motion tracking and task success rates. Furthermore, we show that incorporating trajectories with torque information improves contact force tracking in contact-involved tasks, such as wiping a table. We have successfully transferred our approach to real-world applications. \url{https://opt2skill.github.io}
\end{abstract}


\section{Introduction}
\label{sec:intro}

% \textcolor{red}{[Ye: highlight the keywords of loco-manipulation, full-body trajectory optimization, versatile or versatility, multi-contact behaviors]}

Humanoid robots possess inherent advantages in achieving human-like behaviors~\cite{gu2025humanoidsurvey}. Their anthropomorphic morphology makes them well-suited for human-centered environments and offers the potential to learn from human loco-manipulation behaviors, such as transporting bulky objects, making safe contact with the environment, and performing agile skills such as jumping. Controlling humanoid robots, however, still faces challenges due to their high-dimensional, underactuated nature and the complexities of contact-rich interactions. 

Model-based methods that rely on an accurate model for control have been foundational in enabling humanoid skills. However, their reliability in multi-contact scenarios often limits them to behaviors with simple, pre-defined contacts. In constract, model-free RL that trains skills from scratch achieves robustness to complex contact modes even with environment uncertainties. However, it requires heavy reward tuning to regulate the policy behavior and to cross the sim-to-real gap. As a result, providing RL with a reference has become a mainstream approach. For example, learning loco-manipulation by imitating human motions has shown success on humanoid hardware~\cite{DeepMimic2018, cheng2024expressive, he2024learning, radosavovic2024humanoid}. However, acquiring such data in high quality and for diverse tasks still poses a challenge. Furthermore, these datasets require motion retargeting,
%, which maps collected motions from a source skeletal model to a target robot model. 
and the retargeted motions may not always be kinematically or dynamically feasible, limiting the robot's ability to imitate the original motions effectively, and potentially losing behavior versatility to some extent. Other than using human data, complementing RL with model-based methods for whole-body control of legged robots is also a growing trend~\cite{jenelten2024dtc}.
Along this line, learning dynamically-feasible whole-body motions from model-based methods to achieve diverse humanoid loco-manipulation tasks % and developing a learning-based whole-body controller 
remains an open problem. 

\begin{figure}[t]
\centering
\includegraphics[width=0.9\linewidth]{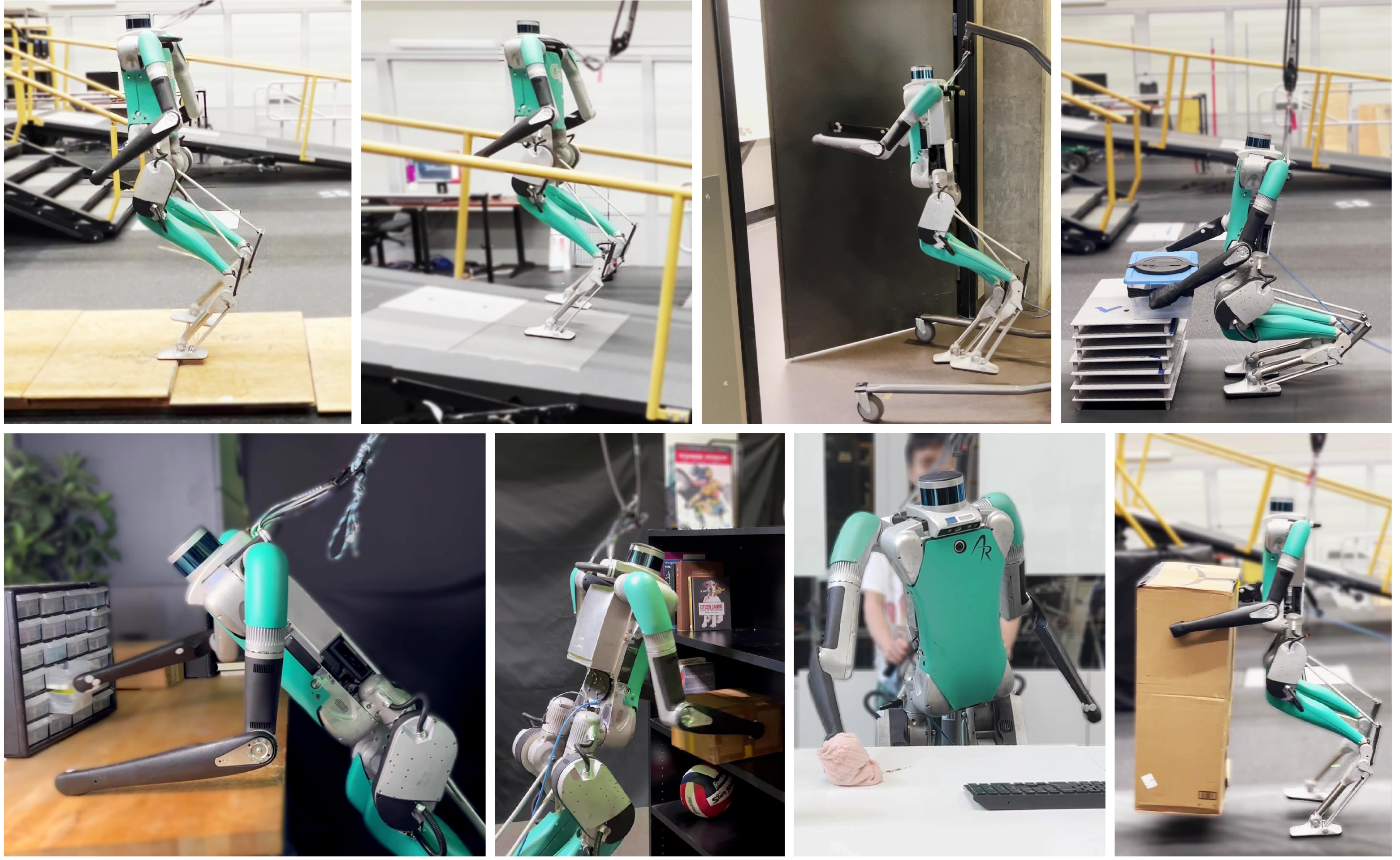}
\caption{The proposed Opt2Skill framework enables a Digit humanoid robot to perform various loco-manipulation tasks by mimicking optimal model-based reference trajectories in real-world scenarios.}
\label{fig:intro}
\vspace{-0.2in}
\end{figure}

 This study presents \textbf{Opt2Skill}, a scalable learning pipeline that transforms whole-body reference motions from trajectory optimization (TO) into humanoid loco-manipulation skills via sim-to-real RL. Our approach leverages differential dynamic programming (DDP)~\cite{mayne1966second, DDP_Tassa} to generate whole-body motions that obey the robot's dynamics and task requirements. Using Opt2Skill, We train RL policies to accurately track these optimal and dynamically feasible motions. Opt2Skill enables a diverse set of loco-manipulation tasks, including bulky-object handling, door traversing, desk-object manipulation, and locomotion over uneven terrains such as stairs, as shown in Fig.~\ref{fig:intro}. Our study indicates that Opt2Skill outperforms policies trained with human data or inverse-kinematics-based references in both tracking accuracy and task success rate. We further show that incorporating joint torque information improves contact force tracking in contact-rich scenarios. Finally, Opt2Skill policies transfer successfully to hardware without online trajectory adaptation.
The contributions of this study are listed as:
\begin{itemize}
\item This work represents the first step to adopting full-order-dynamics-based TO to guide RL that achieves humanoid loco-manipulation tasks. 
\item We show that full-body TO provides higher-quality motion data than mocap and IK baselines, leading to better motion tracking accuracy and task success. We further demonstrate that joint torque information—only available from TO—enhances tracking performance in contact-rich scenarios.
\item We demonstrate the capability of our framework via successful sim-to-real transfer across diverse humanoid loco-manipulation tasks, including multi-contact whole-body manipulation and robust locomotion over stairs and in outdoor environments.
\end{itemize}

\section{Related Work}
\label{sec:related_work}

\subsection{Model-Based Trajectory Optimization for Humanoids}
Model-based motion planning has been crucial in advancing the capabilities of dynamic humanoid robots, enabling accurate loco-manipulation behaviors \cite{gu2025humanoidsurvey}. A popular model-based approach formulates motion planning as an optimal control problem (OCP), incorporating adjustable objectives and constraints to generate versatile motions~\cite{wensing2023optimization}.

Model predictive control (MPC) widely uses OCP for motion planning in an online receding horizon fashion. However, due to the computation constraint, MPC for humanoids predominantly uses various simplified models \cite{kajita2003biped, di2018dynamic, zhou2022momentum, gu2024robustlocomotionbylogic}, which result in conservative motions~\cite{kang2023rl+}. MPC that incorporates full-body models \cite{koenemannwhole, neunert2018whole, katayama2022whole, mastalli2022agile, khazoom2024tailoring} or kino-dynamic models \cite{sleiman2021unified,meduri2023biconmp} still require high computational power and precise contact state feedback, which poses challenges for whole-body planning in real-time, contact-rich loco-manipulation tasks. Alternatively, TO can synthesize high-quality trajectories that are tracked by whole-body controllers. However, model-based whole-body controllers are sensitive to uncertainty in robot dynamics and environment contacts.

While model-based optimization approaches offer the advantage of precision, flexibility, and systematic motion generation for legged robots, they remain a challenging problem for humanoid whole-body loco-manipulation tasks due to the robot's complex, high-dimensional dynamics and the contact-rich, versatile nature of these tasks.

\subsection{Reinforcement Learning for Humanoids}
Reinforcement learning (RL) enables real-time whole-body control of humanoid robots, offering robustness to environmental uncertainty~\cite{humanoid_parkour_learning, dao2023simtoreal}. RL policies are trained through direct interactions with the environment, providing a straightforward training paradigm that learns from scratch without requiring pre-planned motions. However, such RL training often suffers from poor data efficiency. Furthermore, policies learned from scratch can produce unnatural motions~\cite{sferrazza2024humanoidbench} and require significant efforts in reward shaping. 

To address these challenges, RL-based motion imitation has been widely explored to enable humanoid loco-manipulation skills~\cite{DeepMimic2018}. Compared with RL learned from scratch, motion imitation simplifies reward tuning and demonstrates improved data efficiency, yet results in natural motions. Therefore, there is a growing trend in complementing learning-based approaches with motion references for whole-body control of legged robots ~\cite{jenelten2024dtc, Lee2024_MIT_LIP_RL}. 

Motion references come from various sources. While such manually designed references can yield compelling behaviors
\cite{li2022learning, dao2023simtoreal, krishna2024ogmp}, they are often task-specific and do not generalize to diverse or versatile motions.
Human motion capture data is another primary source~\cite{gu2025humanoidsurvey},  and has been successfully transferred to humanoid hardware via RL-based motion imitation~\cite{cheng2024expressive, dugar2024learningmultimodalwholebodycontrol}, enabling both versatile teleoperation~\cite{he2024learning} and autonomous skills~\cite{humanplus}.
However, a significant embodiment gap remains between humans and humanoids, necessitating nontrivial motion retargeting to adapt human movements to robot morphology. So far, motion retargeting is largely restricted to upper-body manipulation tasks, and whole-body loco-manipulation retargeting often yields dynamically infeasible
motions that require substantial data post-processing~\cite{he2024learning}.
As such, the embodiment gap may limit the robot’s ability to faithfully imitate original whole-body motions and reduce behavioral versatility. 

To address these challenges, recent studies have begun using model-based TO to generate data for motion imitation. Model-based TO produces high-quality motion data by ensuring that trajectories adhere to robot dynamics and other physical constraints, such as joint torque limits. TO-based motion imitation has been widely employed on quadrupeds \cite{MIMOC, jenelten2024dtc, fuchioka2023opt}. 
For humanoids, various TO models have been leveraged, including single rigid-body model~\cite{Batke_SRBD_RL}, centroidal dynamics model~\cite{LearnCDM2020}, kino-dynamic model \cite{marew2024soccerkicking}, and full-order-dynamics model \cite{GPS, ICAR23_Full_TO_RL}. However, most studies focus on only locomotion~\cite{Lokesh2022_linearpolicy}. Instead, this work focuses particularly on humanoid loco-manipulation. To achieve this, our TO specifies not only the lower-body leg motions but also the upper-body contact sequence for object manipulation using full-order dynamics. Therefore, our reference trajectories are no longer restricted to periodic motions, highlighting the capability of learning versatile behaviors. Moreover, our work is the first to demonstrate hardware success in policies learned from full-order-dynamics-based TO on a humanoid robot. For loco-manipulation tasks, the high fidelity of the full-order model is beneficial as it eliminates retargeting and is easier to obtain compared to teleoperation. Furthermore, full-order motion data includes joint torques, distinguishing it from other sources such as human motion capture or reduced-order motion. We demonstrate that joint torque information is critical for learning high-dimensional contact-rich skills. 

This work adopts a full-order-dynamics-based TO to guide RL. A closely related study is \cite{fuchioka2023opt}, which also uses TO-based torque supervision for RL. However, our approach differs in several important aspects: we focus on whole-body loco-manipulation using a $30$-DOF bipedal humanoid, while \cite{fuchioka2023opt} focuses on quadrupedal locomotion with an $8$-DOF Solo$8$ robot; we generate trajectories using a full-order dynamics model, whereas they use a simplified rigid-body model; and our work includes extensive simulation and hardware evaluations across seven diverse tasks, along with quantitative analysis of sim-to-real transfer, the impact of torque/force reference information, and comparisons to other motion sources such as inverse kinematics and human demonstrations.

\section{Methods}
\label{sec:method}
\begin{figure*}[t!]
\centering
\includegraphics[width=0.99\linewidth]{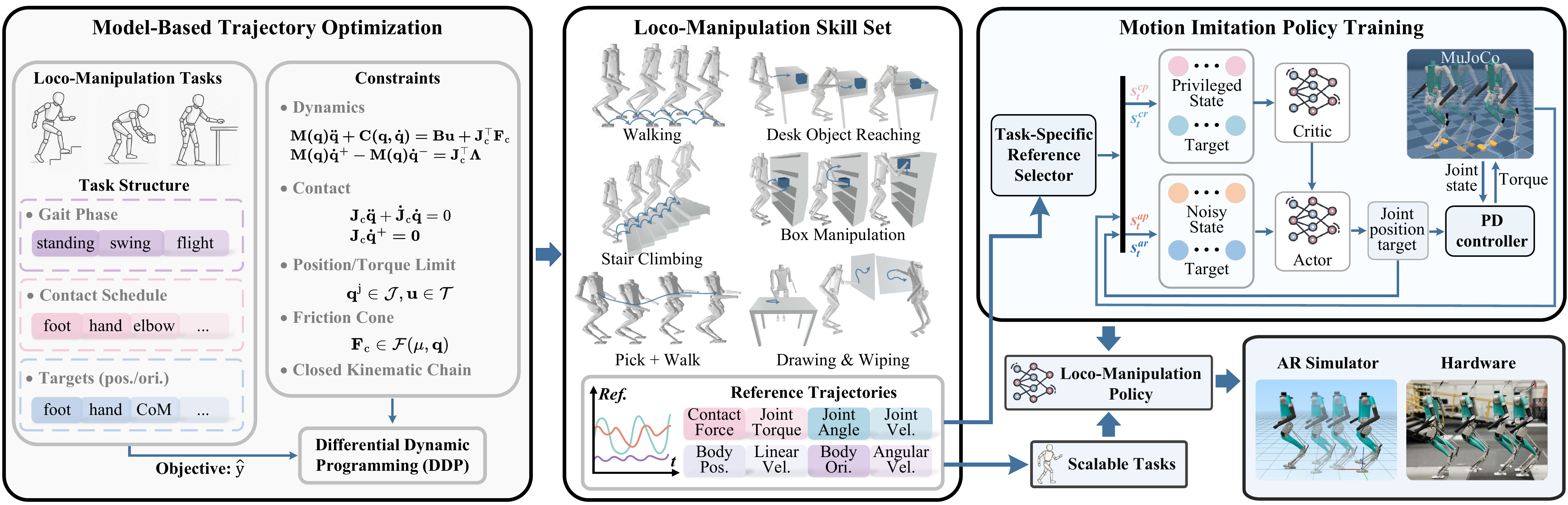}
% \vspace{-0.1in}
\caption{Overall structure of the Opt2Skill framework. (a) We first generate structured, dynamically feasible reference trajectories using trajectory optimization with contact constraints, torque limits, and task-specific objectives. (b) Each trajectory contains joint angles, joint velocities, body position, orientation, linear and angular velocities, and dynamic-relevant quantities such as joint torques and interaction forces. (c) These trajectories serve as supervision signals to train RL policies that predict joint-level targets, tracked by a low-level PD controller. The resulting policies internalize control strategies grounded in model-based optimization while remaining reactive and robust to disturbances, sensor noise, and dynamics variability, enabling direct deployment to real hardware.}
\label{fig:method}
\end{figure*}

Opt2Skill aims to develop an RL-based whole-body controller that enables a humanoid robot to track model-based optimal trajectories. These trajectories contain valuable torque reference that enables a high success rate in multi-contact loco-manipulation tasks. The overall framework is illustrated in Fig.~\ref{fig:method}. We begin by generating dynamically feasible and task-specific motions using DDP in Crocoddyl \cite{mastalli_crocoddyl_2020}. These trajectories serve as high-quality reference motions that encode contact-rich dynamics, torque limits, and task-solving strategies. We generate a diverse set of such trajectories offline and use them to guide the training of RL policies that directly predict joint-level target positions.
By leveraging physically consistent references, the policy learns robust control strategies that track diverse reference motions sampled offline and transfer effectively to real hardware.

\vspace{-0.1in}
\subsection{Whole-Body Trajectory Optimization}
\label{sec:method_TO}
Consider the standard floating-base model of a humanoid robot, with an unactuated $6$-DoF base and a set of $n$-DoF fully-actuated limbs. The equations of motion are given by: 
\begin{equation} \label{eq:fbEoM}
    \begin{aligned}
        &\mathbf{M}(\mathbf{q}) \mathbf{\dot{v}} 
        + \mathbf{C}(\mathbf{q, v}) 
        = \underbrace{ \begin{bmatrix} \textbf{0} \\ \textbf{I} \end{bmatrix} }_{\mathbf{B}} \v u
        + \mathbf{J_{\rm c}^{\top}} \mathbf{F_{\rm c}} &&
        % \vspace{4mm}
        % & \hspace{4cm} \mathbf{\dot{q}} = \mathbf{T}(\mathbf{q})\mathbf{v} &&
    \end{aligned}
    % \vspace{0.5mm}
\end{equation}
where $\mathbf{q} = \begin{bmatrix} \mathbf{q^{\rm b}}; \mathbf{q^{\rm j}} \end{bmatrix} \in \mathbb{R}^{n_{q}}$ and $\mathbf{v} = \begin{bmatrix} \mathbf{v}^{\rm b}; \dot{\mathbf{q}}^{\rm j} \end{bmatrix} \in \mathbb{R}^{n_{v}}$ are the generalized coordinates and velocities, partitioned into base ($\rm b$) and joint ($\rm j$) components. The base coordinates $\mathbf{q^{\rm b}} = \begin{bmatrix} \mathbf{p^{\rm b}};  \vg{\theta}^{\rm b} \end{bmatrix} \in SE(3)$ include base position and orientation represented as unit quaternion, while the base velocities $\mathbf{v}^{\rm b} = \begin{bmatrix} \mathbf{\dot{p}^{\rm b}};  \vg{\omega}^{\rm b} \end{bmatrix} \in se(3) \vspace{1mm}$ are linear and angular velocities in the body frame. $\mathbf{M(q)}$ is the joint-space mass matrix, and $\mathbf{C(q,v)}$ captures the nonlinear effects. $\mathbf{u} \in \mathbb{R}^{n_j}$ is the joint torque command, $\mathbf{J_{\rm c}(q)}$ is the contact Jacobian, and $\mathbf{F}_{\rm c}$ is the stacked contact reaction force vector.

Given this whole-body dynamics, the TO formulation for generating a whole-body trajectory is written as:
\begin{subequations} \label{eq:whole_to}
\begin{align} 
\label{eq:cost}
& 
\underset{
\mathbf{x},
\mathbf{u}
}
{\min} 
& &
\sum_{k=0}^{N - 1} 
\Bigl( 
\| 
\mathbf{y}[k] - 
\hat{\mathbf{y}}[k] \|^2 _Q + 
\| 
\mathbf{u}[k] \|^2_{R} 
\Big) + \\
&
&& 
\| 
\mathbf{y}[N] - 
\hat{\mathbf{y}}[N] \|^2 _{Q_{\rm f}} \nonumber \\
&
\text{subject to} \nonumber \\
\label{eq:hybrid_dynamics}
&
(\textit{\rm Dynamics})         
&& 
\begin{cases} 
\v M(\v q) 
\v {\ddot q} + 
\v C(\v q, \v {\dot q})  = 
\v B 
\v u + 
\mathbf{J_{\rm c}^{\top}} 
\mathbf{F_{\rm c}}\\
\v M(\v q)\v{\dot{q}}^+ - 
\v M(\v q)\v{\dot{q}}^- = 
\mathbf{J_{\rm c}^{\top}} 
\vg 
\Lambda
\end{cases}\\
\label{eq:contact}
&
(\textit{\rm Contact}) 
&& 
\begin{cases}
\v J_{\rm c} 
\v{\ddot{q}}+
\mathbf{\dot{J}_{\rm c}}
\v{\dot{q}} = 
0\\
\textcolor{black}{
\v J_{\rm c} 
\v{\dot{q}}^+ = 
0
}
\end{cases}\\
\label{eq:joint_torque_limits}
&
(\textit{\rm Limits}) 
&& 
\v q^{\rm j} 
\in \mathcal{J}, 
\v u \in \mathcal{T}\\
&
\text{(Friction)} 
&&
\mathbf{F}_{\rm c} 
\in \mathcal{F}(\mu, \mathbf{q}) 
\hspace{1cm} 
\label{eq:friction_cone}
\end{align}
\end{subequations}
where the decision variables include the full-body state vector $\mathbf{x} = \begin{bmatrix} \mathbf{q}; \dot{\mathbf{q}} \end{bmatrix} \in \mathbb{R}^{n_{q}+n_{v}} \vspace{1mm}$ and the joint torque $\v u$. The task space variables $\mathbf{y} = \Phi(\mathbf{q})$ and $\hat{\mathbf{y}}$ are defined based on the specific task. In (\ref{eq:hybrid_dynamics}), the hybrid dynamics constraint includes either the continuous whole-body dynamics (first row), or the impact dynamics (second row), where $\vg \Lambda$ denotes the contact impulse, and $\dot{\v q}^-$ and $\dot{\v q}^+$ are instantaneous velocities before and after the impact. Following (\ref{eq:hybrid_dynamics}), the contact constraints (\ref{eq:contact}) is added for the stance or impact foot assuming a rigid contact, while a zero force is applied to the non-contact foot. Joint limit $\mathcal{J}$, torque limit $\mathcal{T}$, and friction cone $\mathcal{F}$ with coefficient $\mu$ are imposed as constraints in (\ref{eq:joint_torque_limits})-(\ref{eq:friction_cone}).

Considering computational efficiency, we employ a DDP-based method implemented in Crocoddyl~\cite{mastalli_crocoddyl_2020} to solve whole-body optimization. 
To define the desired trajectory $\hat{\mathbf{y}}$ in (\ref{eq:cost}), we use task-specific strategies. For tasks requiring precise foot tracking, such as walking or stair climbing, we specify $\hat{\mathbf{y}}[k]$ at each timestep to enforce desired end-effector trajectories. 
For more flexible behaviors, such as box pickup or desk-object manipulation, we define sparse subgoals and allow the optimizer to generate smooth transitions toward those targets. In all cases, task-specific costs and contact schedules are predefined to produce feasible motions under full-body dynamics. The optimized trajectories include joint and base states, torques, and end-effector motions, which are extracted as reference data for downstream policy training. 

To support robust policy learning across a wide range of loco-manipulation behaviors, we generate reference trajectories for diverse tasks, including \textit{walking}, \textit{stair climbing}, \textit{object pickup}, \textit{drawing}, and \textit{desk-object reaching}. For each task, we systematically vary key motion parameters such as \textit{gait phase}, \textit{contact mode}, \textit{walking speed}, \textit{foot clearance}, \textit{center of mass position}, and \textit{target object location}. Leveraging the flexibility of our TO framework, we can efficiently sample these parameters to produce large datasets of motions. Unlike methods based solely on human demonstrations or heuristic design, our approach generates dynamically feasible trajectories, which contain critical dynamics-relevant properties such as joint torques and interaction forces that are often missing from kinematic demonstrations.  
Detailed task-specific definitions of $\hat{\mathbf{y}}$ and trajectory examples are presented in Sec.~\ref{sec:result}.

\subsection{RL-Based Imitation of Dynamically Feasible Trajectories}
\label{sec:method_RL}

We leverage TO to generate reference motions that satisfy full-body dynamics, contact constraints, and torque limits. These trajectories include joint torques and contact forces, which are critical for learning contact-rich loco-manipulation skills. We incorporate this information directly into the policy observations and reward design, helping the policy learn robust and transferable whole-body behaviors. 

During training, one trajectory is randomly sampled at the start of each episode to initialize the environment and serve as the reference for computing tracking rewards. We train separate policies for each loco-manipulation task, each using its own TO-generated trajectory dataset. At deployment, the policy is provided with partial references from a new offline-generated TO  trajectory tailored to the specific test scenario, without reusing training data or running TO online.

While DDP (used within TO) produces dynamically feasible nominal trajectories, it is rarely employed directly for control due to modeling inaccuracies, contact uncertainties, and high computational cost in real-time settings. To address this, we leverage RL policies that track these trajectories and adapt to real-world disturbances during execution.

\begin{table}[t]
\setlength{\tabcolsep}{1pt}
\renewcommand\arraystretch{1}
\caption{Reward Components and Weights.}
% \scriptsize 
\centering
\label{tab:reward}
\begin{threeparttable}
% \begin{tabular}{l l l l}
\begin{tabularx}{0.95\linewidth} {p{1.3cm} p{2.2cm} X >{\centering\arraybackslash}p{1.5cm}}
\midrule
\textbf{Category} & \textbf{Term} & \textbf{Expression} & \textbf{Weight} \\
\midrule
\multirow{8}{*}{\makecell{Task \\ Reward}}
&Joint Pos. 
& $\exp(-5 \|\mathbf{{\hat{q}}}^{\rm j}_t - \mathbf{{{q}}}^{\rm j}_t\|_2^2)$                 
& $0.30$ \\
& Base Pos. 
& $\exp(-20 \|\mathbf{\hat{p}}^{\rm b}_t - \mathbf{{p}}^{\rm b}_t\|_2^2)$                 
& $0.30$ \\
& Base Ori. 
& $\exp(-50 \|\hat{\vg{\theta}}^{\rm b}_t - {\vg{\theta}}^{\rm b}_t\|_2^2)$       
& $0.30$ \\
& Base Lin. Vel. 
& $\exp(-2 \| \mathbf{\hat{\dot{p}}}_t^{\rm b} - \mathbf{{\dot{p}}}^{\rm b}_t\|_2^2)$             
& $0.30$ \\
& Base Ang. Vel. 
& $\exp(-0.5 \|\hat{\vg{\omega}}^{\rm b}_t - {\vg{\omega}}^{\rm b}_t\|_2^2)$ 
& $0.30$ \\
& End-effector Pos. 
& $\exp(-20 \|\hat{\mathbf{{p}}}^{\rm e}_t - {\mathbf{{p}}}^{\rm e}_t\|_2^2)$         
& $0.30$ \\
& Joint Torque 
& $\exp(-0.01 \|\hat{\mathbf{{u}}}_t - {\mathbf{{u}}}_t\|_2^2)$         
& $0.10$ \\
& Contact Force 
& $\exp(-0.05 \|\hat{\mathbf{{F}}}^{\rm c}_t - {\mathbf{{F}}}^{\rm c}_t\|_1)$         
& $0.10$ \\
\midrule
\multirow{3}{*}{\makecell{Penalty\\ Cost}} 
% & Base Motion 
% & $\| \mathbf{{\dot{p}}}^{\rm b,z}_t\|_2^2 + 0.5\times\| {\vg{\omega}}^{\rm b,xy}_t\|_2^2$           
% & $-1$\\
% & Base Ori. 
% & $\| {\vg{\theta}}^{\rm b, xy}_t\|_2^2$                                   
% & $-2$\\
% & Action Rate 
% & $\|\mathbf{a}_t - \mathbf{a}_{t-1}\|_2^2$                           
% & $-0.3$ \\
& Action Rate 
& $\|\mathbf{a}_t - 2\mathbf{a}_{t-1} + \mathbf{a}_{t-2}\|_2^2$            
& $-0.05$ \\
& Torques 
& $\|\mathbf{u}_t / \mathbf{u}_{\text{limit}}\|_2^2$                                        
& $-0.03$ \\
& Joint Acc. 
& $\|\mathbf{{\ddot{q}}}^{\rm j}_t\|_2^2$                                        
& $-10^{-6}$ \\
\midrule
\end{tabularx}
Note that the ``Base Pos." reward is not included in the ``walking" task.
\end{threeparttable}
\end{table}

\begin{table}[t!]
\centering
\caption{Domain Randomization Parameters.}
\label{tab:domain_randomization}
\renewcommand{\arraystretch}{1.1}
\begin{tabular}{l l l l}
% \begin{tabularx}{0.95\linewidth} {p{1.3cm} p{2.2cm} X >{\centering\arraybackslash}p{1.5cm}}
\toprule
\textbf{Category} & \textbf{Parameter} & \textbf{Type} & \textbf{Range / Std} \\
\midrule
\multirow{5}{*}{Obs.} 
& Joint Pos. 
& Additive (Gauss) 
& $\sigma=0.0875$ \\
& Joint Vel. 
& Additive (Gauss) 
& $\sigma=0.075$ \\
& Base Lin. Vel. 
& Additive (Gauss) 
& $\sigma=0.15$ \\
& Base Ang. Vel. 
& Additive (Gauss) 
& $\sigma=0.15$ \\
& Gravity Proj.  
& Additive (Gauss) 
& $\sigma=0.075$ \\
\midrule
Delays 
% & Action Delay & Uniform & $[0.0, 0.2] \times dt$ \\
& Action Delay 
& Uniform 
& $[0.0, 0.02]\,\text{s}$ \\
\midrule
\multirow{2}{*}{Motor} 
& Motor Strength 
& Scaling (Uniform) 
& $[0.95, 1.05]$ \\
& Kp/Kd Factor 
& Scaling (Uniform) 
& $[0.9, 1.1]$ \\
\midrule
Body 
& Mass 
& Scaling (Uniform) 
& $[0.9, 1.1]$ \\
\midrule
\multirow{3}{*}{Env.} 
& Gravity 
& Scaling (Uniform) 
& $[0.9, 1.1]$ \\
& Friction 
& Scaling (Uniform) 
& $[0.3, 1.0]$ \\
& Terrain 
& Discrete 
& \texttt{flat, rough} \\
\bottomrule
\end{tabular}
% \end{tabularx}
\end{table}

\subsubsection{Problem Formulation} 
We formulate the control problem as an RL problem under a Markov Decision Process (MDP), defined by the tuple $\mathcal{M} = (\mathcal{S}, \mathcal{A}, \mathcal{P}, \mathcal{R}, \gamma)$, with state space $\mathcal{S}$, action space $\mathcal{A}$, transition dynamics $\mathcal{P}$, reward function $\mathcal{R}$, and discount factor $\gamma$. At each timestep $t$, the agent observes a state $s_t \in \mathcal{S}$, selects an action $a_t \sim \pi(a|s_t)$, receives a reward $r_t$, and transitions to a new state $s_{t+1} \sim \mathcal{P}(\cdot|s_t, a_t)$. In real-world deployments, the agent operates under partial observability due to sensor noise and unobservable environmental factors, resulting in a Partially Observable Markov Decision Process (POMDP). The policy $\pi(a|o_t)$ must act based on noisy and incomplete observations $o_t \in \mathcal{O}$, which are incomplete projections of the true state $s_t$. Our goal is to learn an optimal policy $\pi$ that maximizes the expected discounted return $\mathbb{E}[\sum_{t=0}^{T} \gamma^t r_t]$, where the reward encourages accurate tracking of reference trajectories generated via TO.

We train the RL policy in the MuJoCo Playground simulator~\cite{zakka2025mujocoplayground} using an Actor-Critic (AC) architecture optimized by Proximal Policy Optimization (PPO)~\cite{schulman2017proximal}. Both the actor and critic networks are composed of a Multi-Layer Perceptron MLP with hidden sizes of \([512, 512, 256, 256]\). 
Each policy is trained for fewer than $300$ million steps, requiring under $7$ hours on a single NVIDIA RTX$4090$ GPU.

\subsubsection{Imitation Policy}
To enable robust deployment on real hardware, the agent must make decisions based on noisy sensor measurements rather than true state information. To address this challenge, we adopt an asymmetric actor-critic framework~\cite{pinto2017asymmetric}, where the critic has access to privileged state information in simulation, while the actor observes only noisy proprioceptive inputs available during deployment.

The observation space for the critic is defined as
\(
\mathbf{o}_{\text{critic}} = 
\left[\mathbf{s}_t^{cp};\ \mathbf{s}_t^{cr}\right],
\)
where the privileged proprioception \(\mathbf{s}_t^{cp}\) is
\(
[
\mathbf{p}^{b}_t,\
\boldsymbol{\theta}^{b}_t,\
\dot{\mathbf{p}}^{b}_t,\
\boldsymbol{\omega}^{b}_t,\
\mathbf{g}_t,\
\mathbf{q}_{\text{hist}}^{\rm j},\
\dot{\mathbf{q}}^{\rm j}_t,\
\mathbf{p}^{e}_t,\
\mathbf{a}_{\text{hist}},\
\mathbf{{F}}^{\rm c}_t,\
\mathbf{{u}}_t,\
\mathbf{K}_p,\
\mathbf{K}_d
],
\)
which includes the 
humanoid body translation \(\mathbf{p}^{b}_t\), 
orientation \(\boldsymbol{\theta}^{b}_t\), 
linear velocity \(\dot{\mathbf{p}}^{b}_t\), 
angular velocity \(\boldsymbol{\omega}^{b}_t\), 
projected gravity \(\mathbf{g}_t\) (a proxy for base orientation), 
a history of $N = 10$ past motor joint positions sampled every $\delta = 4$ timesteps,
\(
\mathbf{q}_{\text{hist}}^{\rm j} = 
[
\mathbf{q}_{t}^{\rm j},
\mathbf{q}_{t-\delta}^{\rm j},
\ldots,
\mathbf{q}_{t-(N-1)\delta}^{\rm j}
]
\)
(i.e., $50$ Hz sampling from a $200$ Hz control loop), 
motor joint velocities \(\dot{\mathbf{q}}^{\rm j}_t\), 
end-effector position (relative to the torso) \(\mathbf{p}^{e}_t\), 
a history of N past actions sampled with the same schedule,
\(
\mathbf{a}_{\text{hist}} = 
[
\mathbf{a}_{t-1},
\mathbf{a}_{t-(1+\delta)},
\ldots,
\mathbf{a}_{t-(1+(N-1)\delta)}
]
\), contact force \(\mathbf{{F}}^{\rm c}_t\), joint torque \(\mathbf{{u}}_t\),
and the PD gains \(\mathbf{K}_p\), \(\mathbf{K}_d\). 
The reference state \(\mathbf{s}_t^{cr}\) is defined as
\(
[
\hat{\mathbf{p}}^{\rm b}_t,\ 
\hat{\boldsymbol{\theta}}^{\rm b}_t,\ 
\hat{\dot{\mathbf{p}}}^{\rm b}_t,\ 
\hat{\boldsymbol{\omega}}^{\rm b}_t,\ 
\hat{\mathbf{q}}_{t}^{\rm j},\ 
\hat{\dot{\mathbf{q}}}^{\rm j}_t,\ 
\hat{\mathbf{p}}^{\rm e}_t,\
\hat{\mathbf{{F}}}^{\rm c}_t,\
\hat{\mathbf{{u}}}_t
],
\)
where the hat \((\hat{\cdot})\) denotes reference trajectory information.

The observation space for the actor is defined as
\(
\mathbf{o}_{\text{actor}} = 
\left[\mathbf{s}_t^{ap}, \ \mathbf{s}_t^{ar}\right],
\)
where the noisy proprioception \(\mathbf{s}_t^{ap}\) is
\(
[
\tilde{\dot{\mathbf{p}}}^{\rm b}_t,\ 
\tilde{\boldsymbol{\omega}}^{\rm b}_t,\ 
\tilde{\mathbf{g}}_t,\ 
\tilde{\mathbf{q}}_{\text{hist}}^{\rm j},\ 
\tilde{\dot{\mathbf{q}}}^{\rm j}_t,\ 
\mathbf{a}_{\text{hist}}
],
\)
where the tilde \((\tilde{\cdot})\) indicates noisy sensor measurements. The actor also receives partial reference information \(\mathbf{s}_t^{ar}\) =
\(
[
\hat{\dot{\mathbf{p}}}^{\rm b}_t,\ 
\hat{\boldsymbol{\omega}}^{\rm  b}_t,\ 
\hat{\mathbf{q}}_t^{\rm j},\
\hat{\mathbf{{F}}}^{\rm c}_t,\
\hat{\mathbf{{u}}}_t
].
\)
The actor receives partial references to ensure generalization and avoid reliance on noisy or redundant inputs such as global base states or end-effector poses, while the critic has access to full privileged information for stable value estimation.

The action space of the control policy represents offsets from a default standing pose to specify target positions for the $20$ actuated joints. These targets are fed into a PD joint torque controller, which computes torque as
\(
\mathbf{u}_t = 
\mathbf{K}_p 
(
\mathbf{a}_t + 
\mathbf{{{q}}}^{\rm j}_{\text{dflt}} -
\mathbf{q}^{\rm j}_t) - 
\mathbf{K}_d \mathbf{{\dot{q}}}^{\rm j}_t
\).
Here, \(\mathbf{{{q}}}^{\rm j}_{\text{dflt}}\) denotes the default standing joint positions, and $\mathbf{{{q}}}^{\rm j}_t$ and $\mathbf{{\dot{q}}}^{\rm j}_t$ are the measured joint positions and velocities. Our control policy runs at $200$ Hz, while the internal PD controller operates at $1$ kHz in simulation (during training) and $2$ kHz on hardware (during deployment). No explicit filtering is applied between the policy output and the PD control loop. In hardware, we use the same PD gains as in simulation and send torque commands directly to the robot.

\begin{table*}[t]
\setlength{\tabcolsep}{2pt}
\renewcommand\arraystretch{1.2}
\caption{\label{tab:baseline} Tracking performance of Opt2Skill and baselines (Mean $\pm$ SE; lower is better).}
\centering
\scriptsize 
\begin{threeparttable}
\begin{tabular}{llcccccccc}
\toprule
\textbf{Category} & \textbf{Method} & 
$E_{\text{joint}}$ (rad) $\downarrow$ & 
$E_{\text{vel-x}}$ (cm/s) $\downarrow$ & 
$E_{\text{vel-yaw}}$ (rad/s) $\downarrow$ & 
$E_{\text{ee-hand}}$ (cm) $\downarrow$ & 
$E_{\text{ee-foot}}$ (cm) $\downarrow$ & 
$E_{\text{pos-x}}$ ($\%$) $\downarrow$ & 
$E_{\text{pos-y}}$ (mm/step) $\downarrow$ & 
$E_{\text{yaw}}$ (mrad/step) $\downarrow$ \\
\midrule
\multirow{3}{*}{Data} 
& Opt2Skill (Ours) & 
$0.03 \pm 0.01$ & 
$\mathbf{4.86 \pm 4.18}$ & 
$\mathbf{0.06 \pm 0.04}$ & 
$\mathbf{2.00 \pm 0.76}$ & 
$\mathbf{5.23 \pm 1.75}$ & 
$\mathbf{1.14 \pm 0.61}$ & 
$\mathbf{0.06 \pm 0.03}$ & 
$\mathbf{0.02 \pm 0.01}$ \\
& human-data & 
$0.04 \pm 0.03$ & 
$6.68 \pm 8.96$ & 
$0.06 \pm 0.08$ & 
$4.25 \pm 3.49$ & 
$9.42 \pm 8.19$ & 
$4.49 \pm 7.35$ & 
$0.11 \pm 0.18$ & 
$0.04 \pm 0.04$ \\
& IK-data & 
$\mathbf{0.02 \pm 0.01}$ & 
$7.10 \pm 5.73$ & 
$0.06 \pm 0.05$ & 
$5.47 \pm 2.58$ & 
$5.24 \pm 1.54$ & 
$2.09 \pm 0.27$ & 
$0.10 \pm 0.02$ & 
$0.06 \pm 0.01$ \\
\midrule
\multirow{4}{*}{Ablation} 
& Opt2Skill-hist$15$ &
$\mathbf{0.03 \pm 0.01}$ & 
$5.93 \pm 4.81$ & 
$\mathbf{0.05 \pm 0.04}$ & 
$3.11 \pm 1.58$ & 
$6.48 \pm 2.32$ & 
$2.49 \pm 1.40$ & 
$0.12 \pm 0.06$ & 
$0.05 \pm 0.01$ \\
& Opt2Skill-hist$10$ & 
$\mathbf{0.03 \pm 0.01}$ & 
$\mathbf{4.86 \pm 4.18}$ & 
$0.06 \pm 0.04$ & 
$\mathbf{2.00 \pm 0.76}$ & 
$\mathbf{5.23 \pm 1.75}$ & 
$1.14 \pm 0.61$ & 
$\mathbf{0.06 \pm 0.03}$ & 
$\mathbf{0.02 \pm 0.01}$ \\
& Opt2Skill-hist$5$ &
$0.04 \pm 0.01$ & 
$6.46 \pm 4.85$ & 
$0.06 \pm 0.05$ & 
$5.63 \pm 3.54$ & 
$8.59 \pm 4.24$ & 
$\mathbf{1.12 \pm 0.96}$ & 
$0.06 \pm 0.05$ & 
$0.07 \pm 0.01$ \\
& Opt2Skill-hist$0$ &
$0.04 \pm 0.02$ & 
$7.82 \pm 5.46$ & 
$0.10 \pm 0.09$ & 
$6.68 \pm 4.20$ & 
$8.31 \pm 4.60$ & 
$1.32 \pm 0.66$ & 
$0.07 \pm 0.04$ & 
$0.07 \pm 0.02$ \\
\bottomrule
\end{tabular}
\end{threeparttable}
\end{table*}

\subsubsection{Reward Functions, Domain Randomization, and Curriculum}
We design reward functions to guide the robot to track TO-based reference motions and accomplish diverse tasks across various scenarios. The rewards consist of two parts (inspired by ~\cite{DeepMimic2018,radosavovic2024real}): task and regularization rewards, as detailed in Table~\ref{tab:reward}. The task rewards guide the humanoid to track commands such as target velocity, end-effector positions, pickup targets, and contact events for loco-manipulation. Our approach leverages the advantages of the Crocoddyl TO solver to explicitly generate parameterized foot trajectories with controllable step height and timing. These structured, dynamically feasible references improve end-effector tracking and eliminate the need for heuristic reward shaping.

In addition, task rewards such as joint torque and contact force tracking improve the policy's ability to maintain stable contact and adapt to variations in contact conditions during contact-rich loco-manipulation. Regularization rewards promote smooth, physically plausible, and energy-efficient motions by penalizing large torques and abrupt actions, improving robustness and facilitating sim-to-real transfer.

To facilitate sim-to-real transfer, we apply domain randomization during training, including robot dynamics, observation noise, control delays, and external disturbances, inspired by~\cite{radosavovic2024real} and summarized in Table~\ref{tab:domain_randomization}. We also employ a noise curriculum that gradually increases observation noise and penalty strength. This helps the policy learn more stably in early stages and improves generalization in later stages.

\section{Results}
\label{sec:result}
\subsection{Robot System}
Digit is a general-purpose bipedal humanoid robot developed by Agility Robotics. It weighs approximately $48$ kg and has $30$ degrees of freedom (DoF), including $20$ actuated joints—four in each arm and six in each leg. Passive joints in the ankle and shin are connected via leaf spring four-bar linkages. The toe joints (pitch and roll) are indirectly actuated through rods driven by two motors. We use a high-fidelity physics-based simulator from Agility Robotics to evaluate the trained policies before hardware deployment. The simulator provides realistic dynamics and physical properties of Digit and shares many features with the hardware. 

\subsection{Comparison with Different Datasets}
\label{sec:datasets}

To evaluate the effectiveness of Opt2Skill, we compare our TO method against two alternative motion sources:

\begin{itemize}
    \item \textit{Human-retargeted motions (Human)} Obtained by extracting end-effectors and torso motions from human demonstrations, with joint angles obtained through Quadratic Programming (QP)-based IK.
    \item \textit{IK-based trajectories (IK):} Generated from hand-designed keypoint sequences (e.g., footstep and torso motion), converted to joint angles via QP-based IK.
\end{itemize}

\subsubsection{Dataset Generation}
The human-retargeted dataset is selected from the AMASS~\cite{mahmood2019amass} motion capture dataset, where we choose approximately $70$ trajectories containing non-aggressive forward walking motions. Before applying IK, we calibrate the body shape parameters to align with the morphology of Digit. In addition, we scale the torso speeds to match the new leg length. The torso speed ranges from $0$ to $1.5$ m/s, with foot clearance between $10$ and $20$ cm. 

To ensure a fair comparison, both IK and TO datasets are generated with matched speed and clearance ranges. The IK dataset is created by specifying target keypoints (feet, hand, torso) and solving IK at each timestep. The TO dataset uses the same target keypoints but solves a whole-body trajectory using DDP. Both the IK and TO datasets contain $2000$ trajectories, all generated on flat ground. While the human-retargeted dataset is smaller due to difficulty in generating high-quality retargeted trajectories for the robot, the range of motion parameters is matched across datasets. 

\subsubsection{Evaluation Metrics} We evaluate both \textit{per-frame tracking} and \textit{long-term drift}. Per-frame metrics include the mean joint angle error $E_{\text{joint}}$ (rad), base linear velocity error $E_{\text{vel-x}}$ (cm/s), and yaw angular velocity error $E_{\text{vel-yaw}}$ (rad/s). We also report hand and foot end-effector tracking errors $E_{\text{ee-hand}}$ and $E_{\text{ee-foot}}$ (cm), which reflect local task-space accuracy. To assess long-term consistency, we report global drift metrics: the normalized final base position drift $E_{\text{pos-x}}$ ($\%$) along the walking direction relative to the reference path length, and the average per-step drift in lateral base position $E_{\text{pos-y}}$ (mm/step) and yaw orientation $E_{\text{yaw}}$ (mrad/step), which reflect accumulated deviations over time. 

Additionally, to evaluate the adaptability of the learned policies and datasets to rough terrain, we compare their success rates on two challenging scenarios: stairs with varying heights and slopes with increasing inclinations.

\begin{figure}[t!]
\centering
\includegraphics[width=0.95\linewidth]{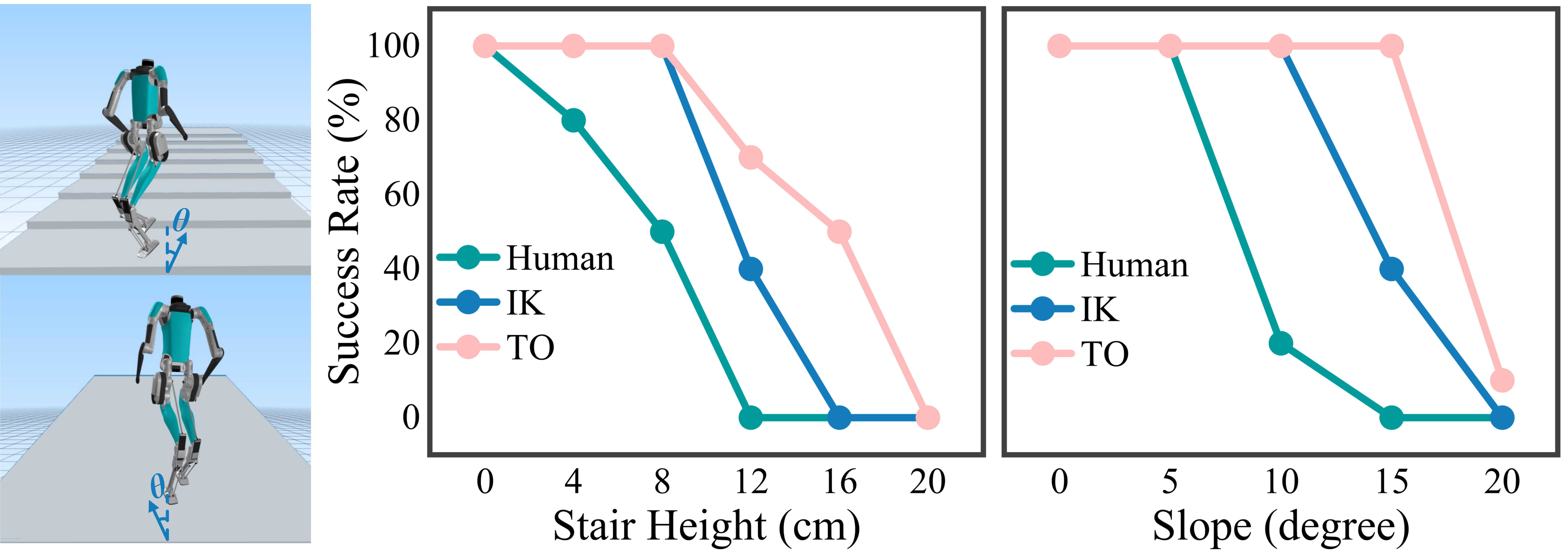}
\caption{Success rate comparison across different terrains. Policies trained on human-retargeted, IK-based, and TO-generated trajectories (with matched parameters) are evaluated on stair heights (left) and slope angles (right). 
}
\label{fig:terrain_baseline}
\end{figure}

\subsubsection{Tracking Performance} Table~\ref{tab:baseline} presents the tracking performance of policies trained with different reference datasets. We evaluate $20$ trajectories per policy. TO and IK trajectories are approximately $14$ seconds long, while human trajectories range from $5$–$6$ seconds. The reported metrics are averaged over the full trajectory duration. Overall, Opt2Skill achieves the best tracking accuracy across most metrics.

Specifically, Opt2Skill achives an average hand tracking error of $2.00$ cm and foot tracking error of $5.23$ cm. In contrast, the human-retargeted policy shows higher errors of $4.25$ cm (hand) and $9.42$ cm (foot), while the IK-based policy yields $5.47$ cm (hand) and $5.24$ cm (foot). Regarding long-term consistency, Opt2Skill achieves the lowest global drift: $1.14\%$ in base position, $0.06$ mm/frame laterally, and $0.02$ mrad/frame in yaw, clearly outperforming the human-retargeted and IK-based policies. Although the IK-based policy achieves the lowest joint angle error ($E_{\text{joint}}=0.02$ rad), its higher task-space tracking errors indicate a fundamental discrepancy between kinematic feasibility and dynamic consistency. These results show that physical feasibility and task adaptability of TO-based references contribute to both local tracking accuracy and long-term global stability.
A Wilcoxon signed-rank test confirmed statistically significant differences in end-effector tracking performance between TO and IK/human datasets ($p<0.001$ for both comparisons).

We further analyze Opt2Skill’s performance under different observation history lengths ($0$, $5$, $10$, and $15$ steps), as shown in Table~\ref{tab:baseline}. The results indicate that incorporating a short to moderate observation history significantly improves tracking accuracy, while overly long histories can deteriorate policy performance. Overall, Opt2Skill with a 10-step history achieves the best overall accuracy.

\subsubsection{Task Success on Rough Terrain}
Fig.~\ref{fig:terrain_baseline} shows success rates of the three policies across varying stair heights and slope angles in the Agility simulator. For each policy and each terrain setting, we conduct $10$ trials with difference initial yaw angles uniformly sampled from $-25^\circ$ to $25^\circ$ to evaluate robustness across diverse orientations ($\theta$ in Fig.~\ref{fig:terrain_baseline}).
Using similar trajectory parameters (base speed $\sim1$~m/s, foot clearance $\sim0.2$~m), Opt2Skill consistently outperforms the human-retargeted and IK-based baselines. Specifically, for stair climbing (Fig.~\ref{fig:terrain_baseline}, left), it maintains high success rates up to $12$~cm steps, while the other policies drop sharply beyond $8$~cm. Similarly, for slope traversal (Fig.~\ref{fig:terrain_baseline}, right), Opt2Skill shows higher robustness at steeper inclinations.

These results demonstrate that Opt2Skill-trained policies, even without any task-specific test-time tuning, exhibit strong robustness to varying terrain. This highlights the advantages of training with offline-generated, dynamically feasible trajectories compared to fixed human or purely kinematic references.

\subsection{Evaluating the Role of Torque Information in Contact-Rich Loco-Manipulation}
\label{sec:contactforce}
We conduct additional simulation experiments to investigate the effect of torque information in contact-rich tasks. Specifically, we focus on a wiping task (see Fig.~\ref{fig:method}) that requires controlled contact force between the end effector and a desk surface. We evaluate four ablation baselines:
\begin{itemize}
    \item \textbf{Pos} that tracks end-effector positions without any contact or torque information in the observation or reward.
    \item \textbf{Pos+F} that adds reference contact force to the observation and includes a contact force tracking reward.
    \item \textbf{Pos+T} that adds reference joint torques to the observation and includes a torque tracking reward.
    \item \textbf{Pos+F+T} that adds both reference contact force and joint torques in the observation, along with corresponding force and torque tracking rewards.
\end{itemize}   
We train all policies using $1400$ TO-generated trajectories with varying desk heights ($0.85$–$0.95$ m) and contact normal forces ($0$–$20$ N).

\begin{table}[t]
\setlength{\tabcolsep}{2pt}
\renewcommand\arraystretch{1.2}
\caption{\label{tab:contactforce} Tracking errors under different contact force references.}
\centering
\begin{threeparttable}
\begin{tabular}{cccccc}
\toprule
\textbf{Category} & \textbf{Ref. Force} & 
\textbf{Pos} & 
\textbf{Pos+F} & 
\textbf{Pos+T} & 
\textbf{Pos+F+T}  \\
\midrule
\multirow{6}{*}{\makecell{Force Err. \\(N)}}
& $0$ & 
$1.2 \pm 2.1$ & 
$\mathbf{0.0 \pm 0.0}$ & 
$\mathbf{0.0 \pm 0.0}$ & 
$\mathbf{0.0 \pm 0.0}$ \\
& $5$ & 
$4.1 \pm 1.7$ & 
$5.0 \pm 0.4$ & 
$3.6 \pm 1.7$ & 
$\mathbf{1.1 \pm 1.4}$ \\
& $10$ & 
$8.7 \pm 2.2$ & 
$3.7 \pm 2.9$ & 
$3.0 \pm 2.6$ & 
$\mathbf{1.5 \pm 2.2}$ \\
& $15$ & 
$13.4 \pm 2.5$ & 
$2.3 \pm 2.4$ &
$2.9 \pm 3.1$ & 
$\mathbf{1.2 \pm 2.7}$ \\
& $20$ & 
$18.7 \pm 2.4$ & 
$3.3 \pm 3.7$ & 
$2.9 \pm 3.8$ & 
$\mathbf{1.6 \pm 3.4}$ \\
& $25$ & 
$23.9 \pm 2.4$ & 
$2.8 \pm 4.3$ & 
$3.0 \pm 4.5$ & 
$\mathbf{2.0 \pm 3.7}$ \\
& Avg. & 
$11.7 \pm 8.3$ & 
$2.8 \pm 3.2$ & 
$2.7 \pm 3.2$ & 
$\mathbf{1.5 \pm 2.6}$ \\
\midrule
\multirow{6}{*}{\makecell{Pos. Err. \\ (cm)}}
& $0$ & 
$\mathbf{2.4 \pm 1.7}$ & 
$3.8 \pm 2.4$ & 
$2.9 \pm 1.5$ & 
$2.6 \pm 0.8$ \\
& $5$ & 
$\mathbf{2.4 \pm 1.7}$ & 
$3.0 \pm 1.5$ & 
$2.9 \pm 1.6$ & 
$2.6 \pm 0.9$ \\
& $10$ & 
$\mathbf{2.4 \pm 1.7}$ & 
$3.0 \pm 1.3$ & 
$2.8 \pm 1.7$ & 
$3.2 \pm 1.4$ \\
& $15$ & 
$\mathbf{2.4 \pm 1.7}$ & 
$3.2 \pm 1.4$ & 
$2.9 \pm 1.7$ & 
$3.2 \pm 1.5$ \\
& $20$ & 
$\mathbf{2.4 \pm 1.7}$ & 
$3.8 \pm 1.5$ & 
$3.0 \pm 1.6$ & 
$2.9 \pm 1.3$ \\
& $25$ & 
$\mathbf{2.5 \pm 1.7}$ & 
$4.6 \pm 1.7$ & 
$3.0 \pm 1.6$ & 
$2.6 \pm 1.4$ \\
& Avg. & 
$\mathbf{2.4 \pm 1.7}$ & 
$3.5 \pm 1.8$ & 
$2.9 \pm 1.6$ & 
$2.9 \pm 1.3$ \\
\bottomrule
\end{tabular}
\textbf{Top:} Contact force tracking error in Newtons. \textbf{Bottom:} Right hand position tracking error in centimeters. Results are averaged over $10$ TO-generated trajectories per group with randomized desk heights (range: $0.85$–$0.95$ m) for a wiping task.
\end{threeparttable}
\end{table}

\begin{figure}[t!]
\centering
\includegraphics[width=0.95\linewidth]{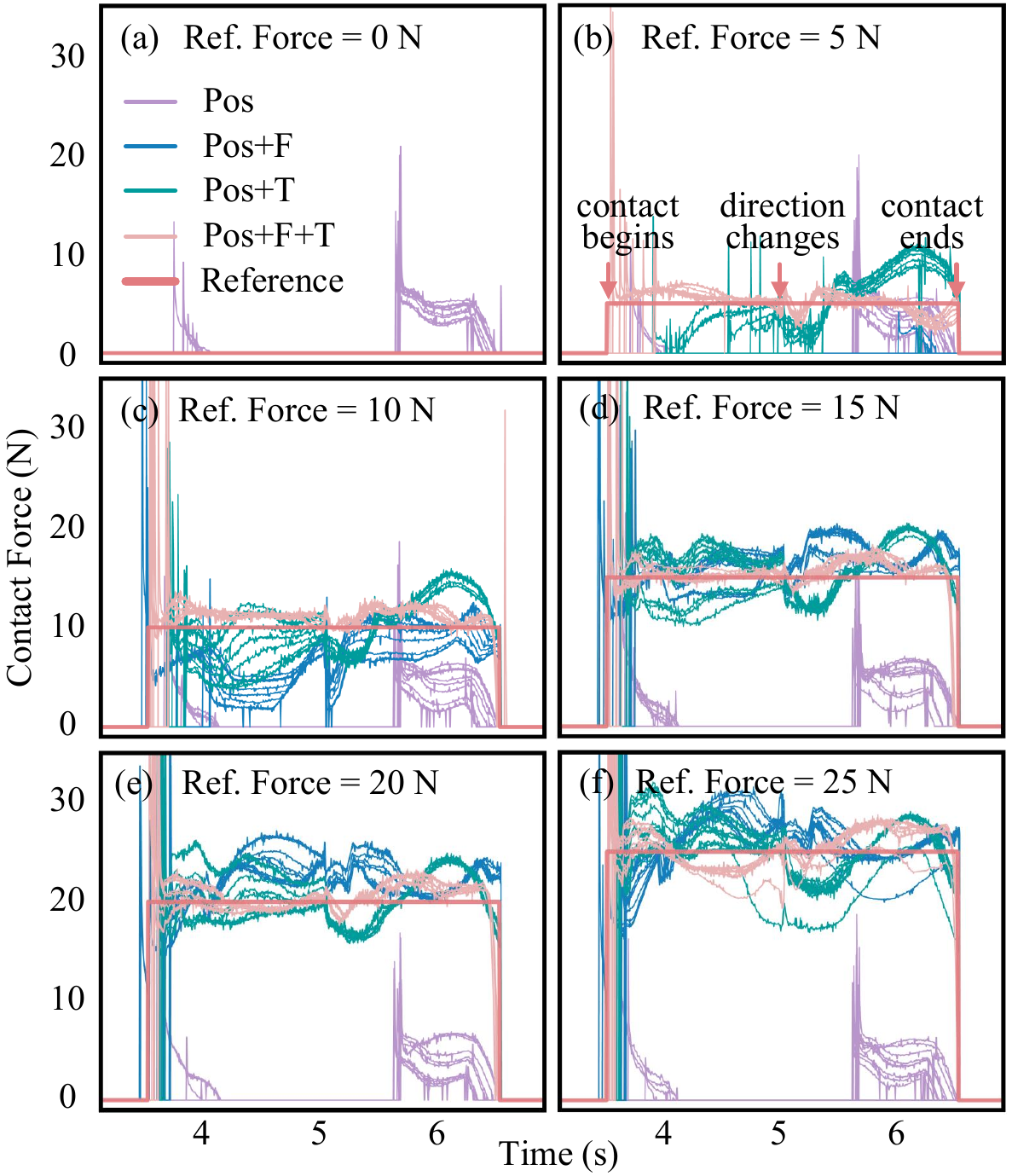}
\caption{Contact force profiles across all four policies under varying reference force levels. Each subplot shows $10$ trajectories per policy.}
\label{fig:contactforce}
\end{figure}

\subsubsection{Performance Analysis}
To evaluate the effectiveness of each policy, we conduct six groups of experiments with different reference contact forces levels, ranging from $0$ to $25$~N. Each group consists of $10$ TO-generated trajectories with randomized desk heights.

\begin{figure*}[t]
\centering
\includegraphics[width=0.97\linewidth]{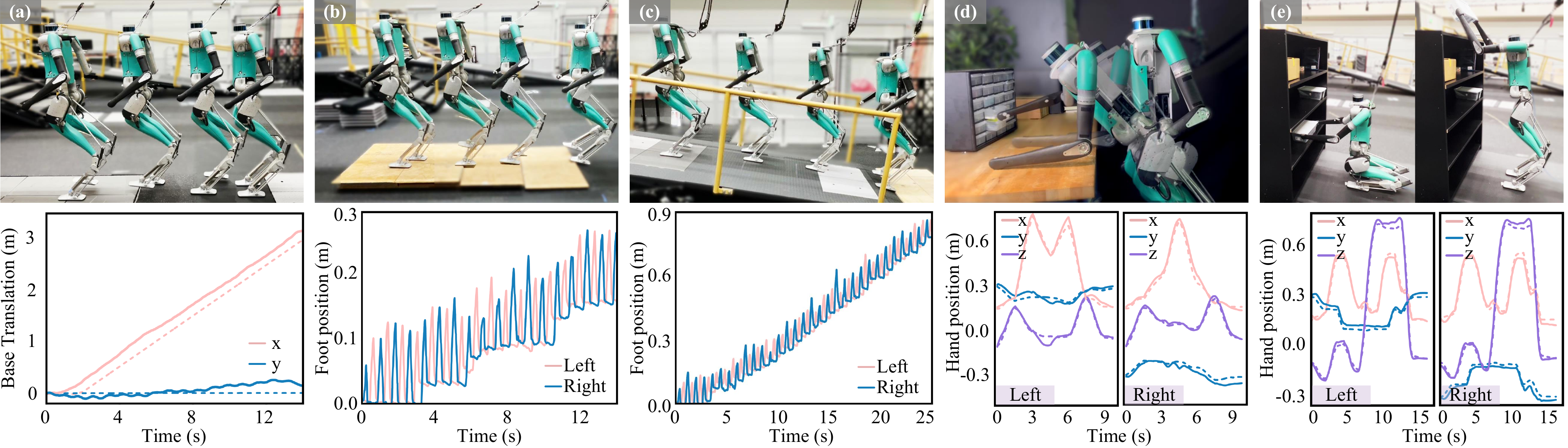}
\caption{Snapshots of hardware experiments with data plots. (a) Flat-ground locomotion with accurate base position tracking. (b-c) Walking up a stair and a ramp  ($\sim19.5^\circ$) with foot $z$-position plots indicating elevation. (d) Desk object reaching, with plots of end-effector tracking. (e) Box pick-up from one shelf layer to another. Note that in (d) and (e), end-effector positions are expressed relative to the root. Dashed lines indicate the reference trajectories.}
\label{fig:real}
\end{figure*}

\begin{figure}[t]
\centering
\includegraphics[width=0.95\linewidth]{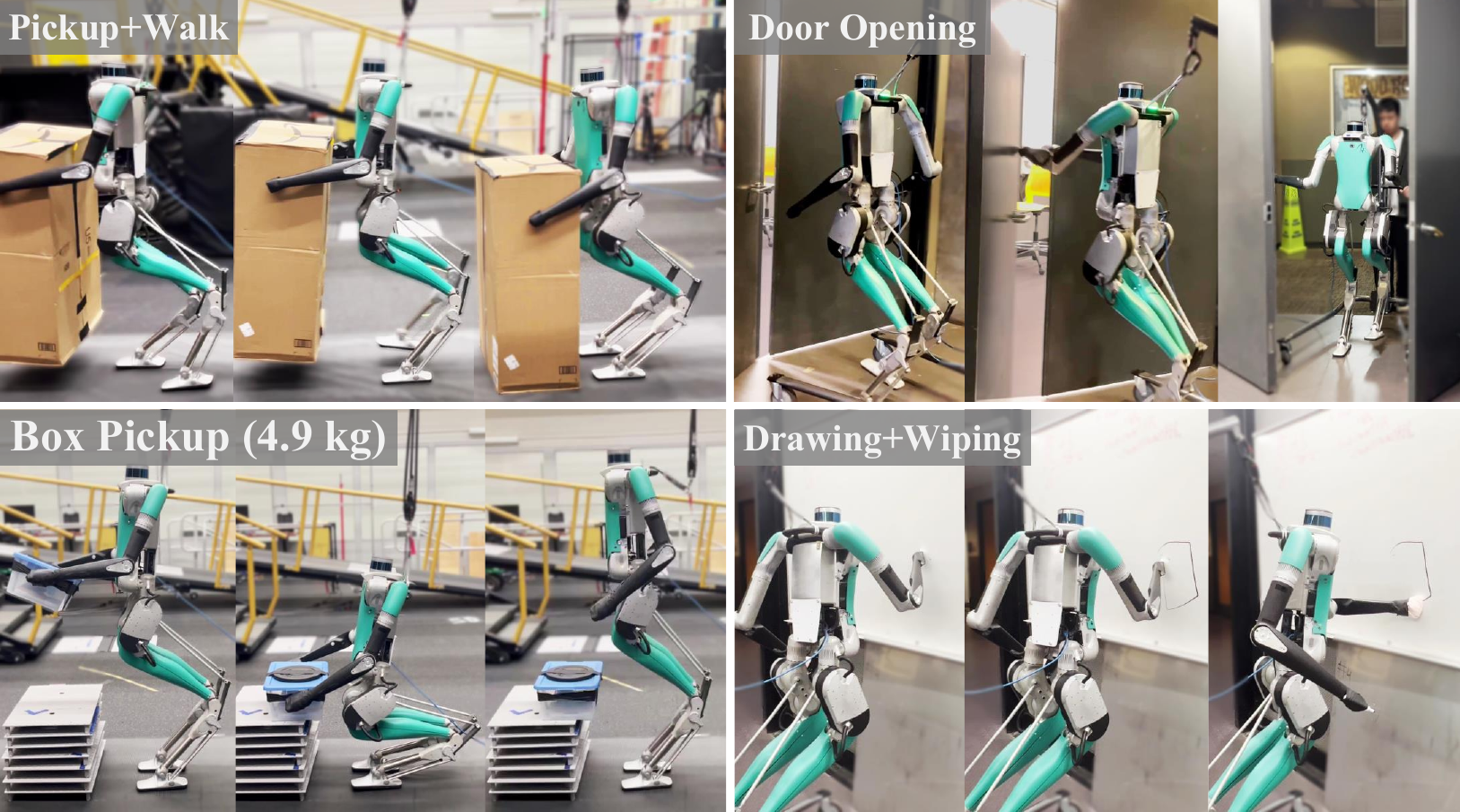}
\caption{Snapshots of hardware experiments demonstrating the \textit{pickup+walk}, \textit{door opening}, \textit{box pickup} and \textit{drawing+wiping} tasks.}
\label{fig:hardware_new}
\vspace{-0.1in}
\end{figure}

Table.~\ref{tab:contactforce} shows the contact force and right end effector position tracking errors across varying contact force levels. The \textbf{Pos} policy achieves the best position tracking across all scenarios. This is expected, as it optimizes soley for position accuracy without being constrained by contact  or torque objectives. However, it produces similar force patterns regardless of the reference force and either fails to generate sufficient contact force (Figs.~\ref{fig:contactforce}(b)--(f)) or overshoots when no force is required (Fig.~\ref{fig:contactforce}(a)), due to lack of contact awareness. 

Adding reference contact force to the observation and reward in \textbf{Pos+F} improves average force tracking accuracy, but instability remains in certain conditions. For instance, in Fig.~\ref{fig:contactforce}(b), \textbf{Pos+F} fails to produce the desired $5$~N contact force. In contrast, \textbf{Pos+T}, which includes reference torques in the observation and reward, shows improved contact force tracking (average error: $2.7$ N) and better position tracking (average error: $2.9$ cm) compared to \textbf{Pos+F} (Table.~\ref{tab:contactforce}). While the reference force in \textbf{Pos+F} helps the policy learn how much force to apply, the reference torque in \textbf{Pos+T} provides dynamically grounded information about \textit{when} and \textit{how} to apply it, resulting in more physically consistent behavior. 

Finally, \textbf{Pos+F+T}, which combines both reference contact force and torque signals, achieves the best overall force tracking (Table.~\ref{tab:contactforce}). Its position tracking is also competitive with \textbf{Pos}, and it exhibits the most stable force application across all scenarios (Fig.~\ref{fig:contactforce}). This highlights the complementary nature of contact force and torque signals, and the critical role of reference torque in enabling generalizable, physically grounded contact behaviors—information uniquely available in TO-generated datasets but absent in IK or human datasets.
Moreover, policies trained with TO-generated references adapt when the environment deviates from nominal settings (e.g., table height). For instance, with a trajectory defined for $0.95$ m height and $15$ N contact force, the policy maintained wiping at $0.9$ m and $1.0$ m table heights, achieving average forces of $9.6\pm1.5$ N and $21.3\pm2.0$ N, and tracking errors of $4.1\pm1.6$ cm and $4.5\pm2.0$ cm, respectively.
A Wilcoxon signed-rank test confirmed statistically significant differences in both contact force and hand position tracking errors between \textbf{Pos+F+T} and other baselines across all force levels ($p < 0.001$). 

The integration of force and torque signals significantly enhances tracking performance, a contribution we believe is essential for high-fidelity imitation of complex dynamic skills. However, we acknowledge this necessitates a more complex pipeline that requires an offline TO stage prior to policy training. We anticipate that future advances in sensing technology and force estimation will simplify the data collection process, further facilitating the adoption of such high-fidelity force signals.

\subsection{Hardware Experiments}
We demonstrate the sim-to-real performance of our Opt2Skill through five hardware experiments. Snapshots of the tasks and their corresponding data plots are shown in Fig.~\ref{fig:real}. 
%\st{For task with defined reference trajectories,} 
We compare the robot's measured states with the corresponding references to evaluate tracking performance. The robot states are obtained from onboard sensors and estimated using the built-in state estimator provided by Agility Robotics.

For locomotion tasks, Fig.~\ref{fig:real}(a) shows accurate base position tracking in the $x$ and $y$ directions during flat ground walking. We generate the reference trajectories by specifying a fixed gait frequency, 
%($2.5$ Hz), 
step length, 
%($-0.5$ to $0.8$ m), 
and foot clearance, 
%($0.1$-$0.2$ m), 
using cubic-spline interpolation to compute desired foot trajectories $\hat{\mathbf{y}}$ described in Sec.~\ref{sec:method_TO}. Fig.~\ref{fig:real}(b) and (c) show the foot height as the robot successfully walks up a stair and a ramp. Notably, Opt2Skill achieves rough terrain walking capability despite the reference being designed only for flat ground scenarios. The robot also maintains toe compliance when stepping on stair edges and ramp slopes—enabled by low PD gains in the low-level controller. 

Fig.~\ref{fig:real}(d) demonstrates a multi-contact whole-body manipulation task, where the robot reaches for an object on a desk located well beyond its support polygon. We generate these reference trajectories via motion phases: 
(1) one arm moves to the target position;
(2) the elbow maintains contact with the desk;
(3) the opposite arm reaches the target position and returns;
(4) release elbow contact; 
(5) return torso and hands to the initial pose. 
These target positions serve as the desired trajectories $\hat{\mathbf{y}}$ in Sec.~\ref{sec:method_TO}. To achieve this, the legs coordinate with upper-body motion to stabilize and enable the forward-reaching behavior, reflecting the necessity of whole-body motion coordination. This example highlights the ability of Opt2Skill to handle high-dimensional loco-manipulation tasks.  In Fig.~\ref{fig:real}(e), the robot squats down to pick up a box from a lower shelf and place it on a higher one. Trajectories are generated by: (1) squat down; (2) both arms reach and squeeze to grasp the box; (3) retract arms slightly and stand up; (4) reach to the target and release; (5) return torso and hands to the initial pose. During training, hand targets range from $[0.40, \pm0.10, 0.55]~\mathrm{m}$ to $[0.50, \pm0.20, 1.70]~\mathrm{m}$. In deployment (Fig.~\ref{fig:real}(e)), the hands reach $0.49$~m (lower) and $1.76$~m (upper), indicating that the policy can handle moderate variations beyond the training range.

The plots in Fig.~\ref{fig:real}(d) and (e) show the tracking performance of both hand positions during these complex motion sequences. In the \textit{desk object pushing} task, Opt2Skill achieves an average tracking error of $3.7$ cm for the left hand and $3.8$ cm for the right. In the \textit{shelf box pick-up} task, the errors are $3.6$ cm for both hands. Notably, a deviation along the $y$-axis occurs when the robot makes contact with the box, caused by a size mismatch between the reference and the real-world box (approximately $2$ cm in the $y$ direction for each hand). 

Additionally, we conduct four more real-world tasks: \textit{heavy box pickup}, \textit{pickup+walk}, \textit{door opening}, and \textit{drawing+wiping}. as shown in Fig.~\ref{fig:hardware_new}. In the first task, the robot squats down and lifts a box weighing up to $4.9$ kg. the trajectory is generated by predefining grasping goals followed by lifting goals for both hands. In the second task, the robot picks up a box from a standing pose and then walks forward with it, using a trajectory obtained by combining the pickup and locomotion tasks. In the third task, the robot pushes open two heavy fire doors using both arms while walking, with end-effector target positions added to the locomotion trajectory.  Finally, the robot uses its right hand to draw various lines on a whiteboard and then wipes them off using its left hand, following a sequence of end-effector subgoals constrained to a planar surface.

These results illustrate that our framework maintains compliant contact, adapts effectively to environmental variations, and enables the robot to perform everyday loco-manipulation behaviors. All policies are trained using over $1000$ reference trajectories per task with randomized parameters, supporting the versatility and generalization capability of the Opt2Skill framework across diverse real-world scenarios.

\section{Conclusion}
\label{sec:conclusion}
In this paper, we present a TO-guided RL pipeline for humanoid loco-manipulation. We show that the RL tracking performance is affected by the quality of the motion reference. The full-body-dynamics-based TO provides high-quality and dynamically-feasible trajectories. Based on such trajectories, motion imitation yields better tracking performance, especially through the use of torque information. We demonstrate our sim-to-real results on the humanoid robot Digit with versatile loco-manipulation skills, including dynamic stair traversing, multi-contact box manipulation, and door traversing.

% \section{Acknowledgment}
% \label{sec:acknowledgment}

\bibliographystyle{IEEEtran}
\bibliography{bibliography}

% \balance

\end{document}